\documentclass[runningheads]{llncs}

\usepackage{graphicx}
\usepackage{color}
\usepackage[hang,small,bf]{caption}
\usepackage[subrefformat=parens]{subcaption}
\usepackage{here}
\usepackage{xspace}
\newcommand{\Ours}{T$^3$\xspace}

\begin{document}
\title{TrueType Transformer: Character and Font Style Recognition in Outline Format}
\titlerunning{TrueType Transformer}
\author{Yusuke Nagata \and
Jinki Otao\and
Daichi Haraguchi\and
Seiichi Uchida\orcidID{0000-0001-8592-7566}}
\authorrunning{Y. Nagata et al.}
\institute{Kyushu University, Fukuoka, Japan}
\maketitle  
\begin{abstract}
We propose TrueType Transformer (\Ours), which can perform character and font style recognition in an outline format. The outline format, such as TrueType, represents each character as a sequence of control points of stroke contours and is frequently used in born-digital documents. \Ours is organized by a deep neural network, so-called Transformer. Transformer is originally proposed for sequential data, such as text, and therefore appropriate for handling the outline data. In other words, \Ours directly accepts the outline data without converting it into a bitmap image. Consequently, \Ours realizes a resolution-independent classification. Moreover, since the locations of the control points represent the fine and local structures of the font style, \Ours is suitable for font style classification, where such structures are very important. In this paper, we experimentally show the applicability of \Ours in character and font style recognition tasks, while observing how the individual control points contribute to classification results.

\keywords{Outline format  \and Font style recognition \and Transformer.}
\end{abstract}
\section{Introduction}\label{sec:intro}
\begin{figure}[t]
    \centering
    \includegraphics[width=0.7\textwidth]{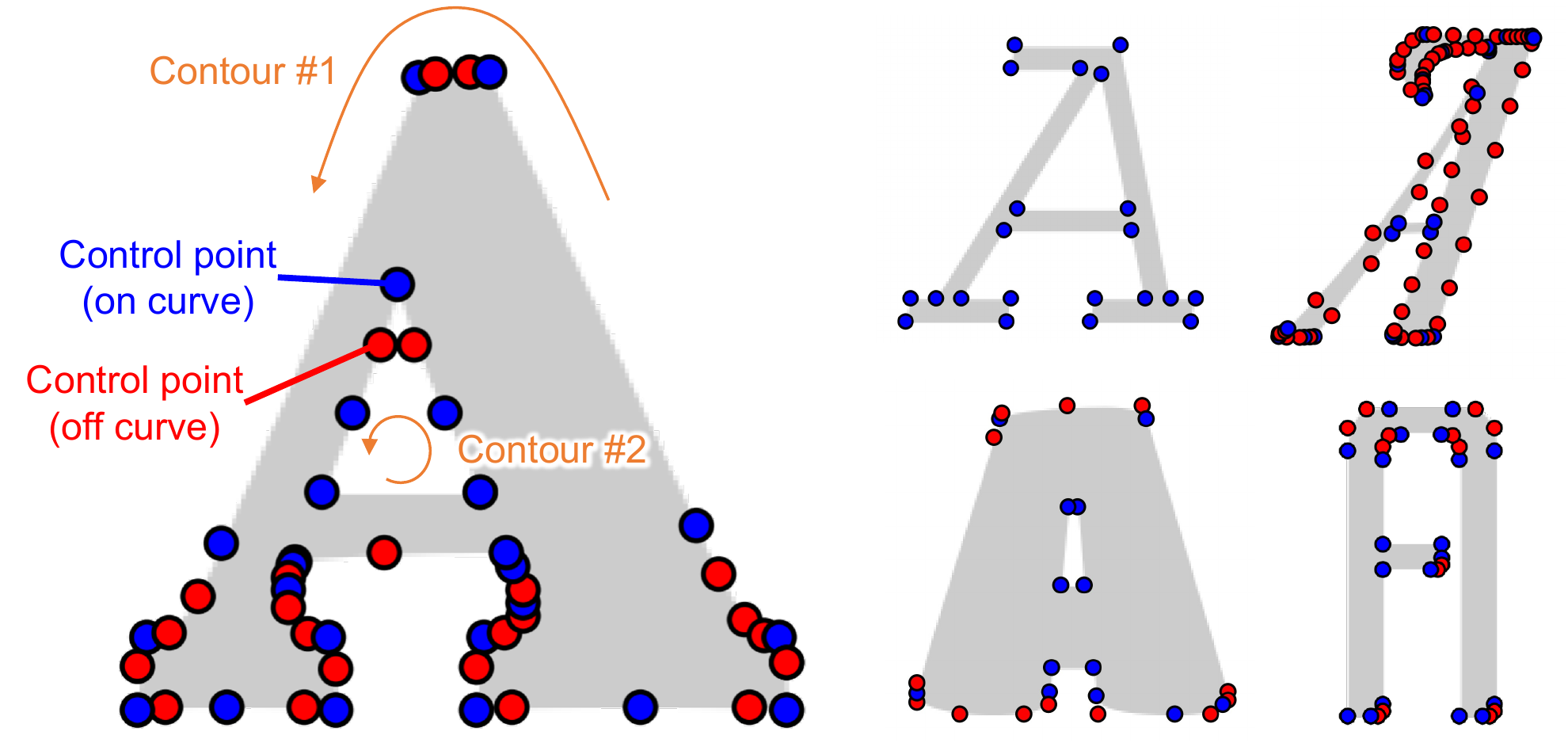}\\[-3mm]
    \caption{An outline format. Each stroke contour is represented as a sequence of {\em control points}. A control point is {\it on curve} or {\it off curve}. In the former case, the stroke outline passes through the control point, whereas the control point controls the contour curvature in the latter case. Roughly speaking, TrueType follows this format.}  
    \label{fig:outline}
\end{figure}

In a born-digital document, text data is embedded into the document as various vector formats, including the {\em outline format}, instead of the bitmap image format and the ASCII format (i.e., character codes plus font names). Fig.~\ref{fig:outline} illustrates the outline format. Each character is represented as single or multiple contours, and each contour is represented by a sequence of {\em control points}. Even for the same letter `A,' its contour shape is largely variable, depending on its typeface (i.e., font style). In other words, the outline format conveys the font style information by itself. Much graphic design software (such as Adobe Illustrator) have a function of converting a character image in a specific font into the outline format.\par
The outline format is often used by graphic design experts because of the following two reasons.

\begin{description}
\item{Reproducibility}: Imagine that an expert designs a web advertisement with texts in a special font X and sends it to another expert who does not have the font X in her/his design environment. If the texts are not in the outline format (i.e., the texts are represented as their character codes plus font names), a different font will be used for showing the texts as the substitution of X. In contrast, if they are in the outline format, exactly the same text appearance is reproduced.  
\item{Flexibility}: Experts often modify font shapes for a specific purpose. Especially, the contour shapes of characters in logotypes and posters are often slightly modified from their original font shape in order to enhance their uniqueness and conspicuousness. Kerning (i.e., space between the adjacent characters) is also tuned by ignoring the default kerning rule of the font. The outline format has the flexibility to represent texts with the unique shapes and kernings.
\end{description} 
\par  

\begin{figure}[t]
\begin{minipage}[b]{0.45\textwidth}
    \centering
    \includegraphics[width=\linewidth]{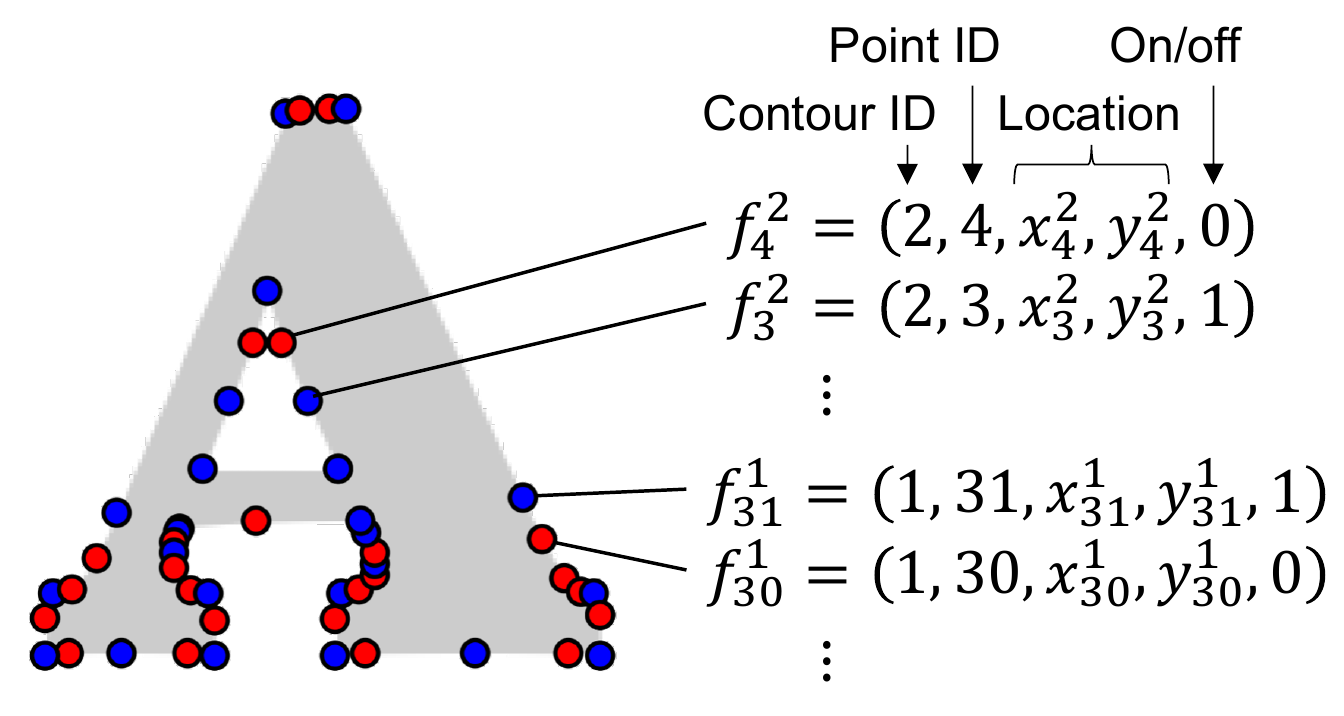}\\
    (a) Control points and their representation
\end{minipage}
 \hspace{0.05\linewidth}
\begin{minipage}[b]{0.45\textwidth}
        \centering
    \includegraphics[width=\linewidth]{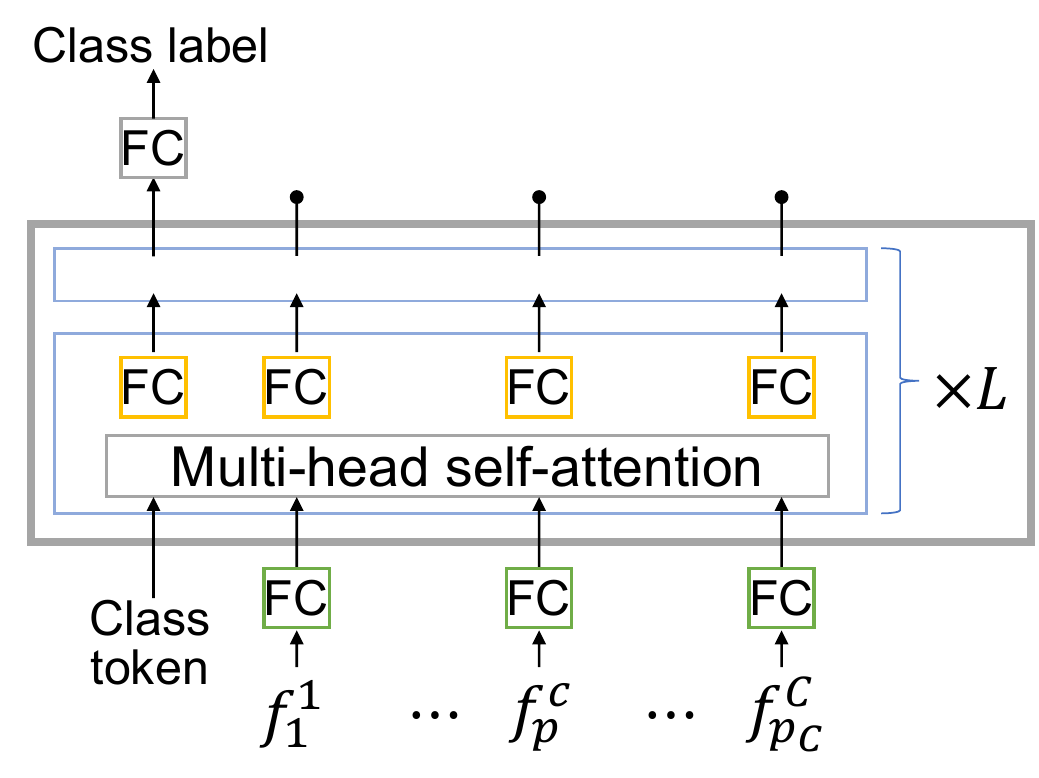}\\
    (b) Transformer (Encoder)
\end{minipage}
\caption{The overview of the proposed TrueType Transformer (\Ours).}
\label{fig:model}
\end{figure}

In this paper, we attempt to recognize characters in their outline format by a Transformer-based neural network, called TrueType Transformer (\Ours), {\em without} converting them into a bitmap image format. Fig.~\ref{fig:model} shows the overview of \Ours. In fact, there are two practical merits to recognizing characters directly in the outline format.
\begin{itemize}
    \item First, we do not need to be careful of the image resolution. If it is too low, the recognition performance becomes poor; if it is too high, we need to waste unnecessary computation costs. Recognizing characters in the outline format is free from this unnecessary hyperparameter, i.e., the resolution.  
    \item The second and more important merit is that the outline format will carry very detailed information about font styles. Font styles are always designed very carefully by font designers, and they pay great attention about the curvatures of contours and fine structures, such as serifs. In the image format, those design elements are often not emphasized. In the outline format, the location and the numbers of control points show the design elements explicitly. Consequently, to understand the font styles, the outline format will be more suitable. 
\end{itemize}
\par

Despite these merits, there are two issues in handling the outline format for recognition tasks. First, characters in the outline format have a different number of control points, and thus their input dimension is also variable. Second, the control points should not be treated as just a point set. For example, suppose we have a character consisting of two contours (such as `A' on the left side of Fig.~\ref{fig:outline}). By cascading the two sequences obtained from both contours, a single control point sequence can be obtained. In this case, in order to understand, for example, the stroke thickness, it is not enough to use the relationship between adjacent points on the sequence; instead, we need to know the positional relationship between points that are spatially close but distant in the sequence. Therefore, the mutual relationships between all the control points are often important to understand the shape in the outline format.
\par

Transformer used in \Ours is very suitable for dealing with the above issues on the recognition tasks with the outline format. For the first issue, Transformer has sufficient flexibility in the input length varieties; in fact, Transformer has been originally developed for natural language texts, which have no fixed length.  For the second issue, the self-attention mechanism inside Transformer can learn the importance of the individual relationships among all inputs, i.e., control points. \par

The purpose of this paper is twofold. The first purpose is to prove that characters are recognizable in the outline format. As noted above, outline-based recognition has the merits for dealing with born-digital documents. To the best knowledge of the authors, this is the first attempt at outline-based character recognition. For this purpose, we consider two different tasks; character class recognition and font style recognition. The former is the orthodox character recognition task and the latter is to classify character data into one of font style classes, such as \textit{Serif} and \textit{Display}. The latter task needs to be more careful of local shapes and thus is more suitable for understanding the merit of the outline-based recognition. \par

The second purpose is to analyze the importance of individual control points for the recognition tasks. This is important for a scientific question to understand discriminative parts for a specific character class or font style class. The analysis is formed by visualizing the self-attention weight, which directly shows the importance of the control points. The method called attention rollout~\cite{abnar2020quantifying} is employed for the visualization. We will observe how the important control points are different between the character recognition task and the font style recognition task. 
\par

We summarize our main contributions as follows.
\begin{itemize}
    \item  To the authors' best knowledge, this is the first attempt to tackle character and font style recognition tasks in an outline format. Recognizing in the outline format is practically useful for dealing with born-digital documents. 
    \item To conduct the above recognition tasks, we propose TrueType Transformer (\Ours) that accepts a set of control points of the character outline as its input. 
    \item \Ours showed higher accuracy than two image-based models (ResNet and ViT~\cite{dosovitskiy2020image}) in the font style recognition task, which needs to be more careful of fine outline shapes. 
    \item We analyze the learned attention weight to understand the various trends of important control points for a specific character class or font style class. The analysis results, for example, show 
    that important control points are very different between the recognition tasks and also among the four style classes.
\end{itemize}

\section{Related work}
\subsection{Transformer}
Transformer~\cite{vaswani2017attention} began to be used and has updated the state-of-the-art records~\cite{devlin2018bert,liu2019roberta} in the field of natural language processing (NLP).
The transformer initially consists of an encoder and decoder and can perform a language translation. One of the main characteristics of the transformer is that it is possible to process sequential data, i.e., textual data.
This characteristic is compatible with our outline data because the outline data is regarded as sequential data as we described in the above Section~\ref{sec:intro}. 
\par

In recent years, the transformer also has been used in the field of computer vision and image classification~\cite{khan2021survey}. Vision Transformer (ViT)~\cite{dosovitskiy2020image} is one of the most famous applications of Transformer for image classification.
Unlike transformer for NLP, Transformer for image classification often accept the constant number of input (because the input image size is fixed), and only a transformer encoder is used.\par

The visualization of attention in Transformer is useful to understand not only the internal behavior of Transformer but also the important elements among all the inputs. Abnar \textit{et al.,}~\cite{abnar2020quantifying} proposed attention rollout to visualize attention weight by integrating the features extracted by each transformer layer.
Chefer \textit{et al.,}~\cite{chefer2021transformer} proposed the visualization method of attention weight based on a specific formulation while maintaining the total relevancy in each layer.
\par
In this paper, we propose \Ours that utilize the characteristics of Transformer to recognize character class and font style from the outline. 
In the experiments, we visualize the attention weights by using the above rollout for understanding important control points (i.e., control points that have a larger attention weight) in character and font style recognition tasks. 

\subsection{Font analysis by using vector format}
To the authors' best knowledge, there is no attempt at character recognition in the outline format. In fact, for many years, the mainstream of font analysis has used bitmap image format (or raster format). 
However, since digital font data is provided in the TrueType format or OpenType format, it is quite natural to deal with them in the vector format. \par
Nowadays, we can find several font-related attempts by using other vector formats, such as scalable vector graphics (SVG). Especially for font generation and conversion tasks, focus on the advantage of vector format because it can be free from resolution or scale. 
Cambell et al.~\cite{Campbell2014} have done pioneering work for outline-based font generation, where a fixed-dimensional font contour model is represented as a low-dimensional manifold. More recently, 
Lopes et al.~\cite{lopes2019learned} proposed a VAE-based model to extract the font style feature from images and SVG decoder for converting the feature into the SVG format.  Carlier et al. proposed DeepSVG~\cite{carlier2020deepsvg}, which is a font generation model that outputs font images in the SVG format.  Im2Vec~\cite{reddy2021im2vec} by
Reddy et al. generate a vector path from an image input. DeepVecFont~\cite{wang2021deepvecfont} by
Wang et al. can address font generation in both image and sequential formats by an encoder-decoder framework.\par 
\par
Unlike the above font generation methods, we mainly focus on recognition tasks by using an outline format with an arbitrary number of control points. For this purpose, we utilize Transformer, which characteristics perfectly fit our problem, as noted in Section~\ref{sec:intro}. Moreover, by using the learned attention in Transformer, we analyze the important control points for character class recognition and font style recognition.

\section{TrueType Transformer (\Ours)}

\subsection{Representation of Outline} 
As shown in Fig.~\ref{fig:model}~(a), each control point is represented as a five-dimensional vector $f^c_p$, where $c\in\{1,\ldots,C\}$  is the contour ID and $C$ is the number of contours in a character data. The point ID $p\in\{1,\ldots,P^c\}$ shows the order in the $c$-th contour\footnote{In the outline format, the start point and the endpoint of each contour are located at exactly the same position, that is, $f^c_1=f^c_{P^C}$.}. Consequently, the character data is represented by a set of $N=\sum_c P^c$ vectors and fed to Transformer in the order of $\mathbf{f}=f^1_1,\ldots,f^1_{P^1},f^2_1,\ldots,f^c_p,\ldots,f^C_{P^C}$.
\par
The five elements of $f^c_p$ are the contour ID $c$, the font ID $p$, the location $(x^c_p,y^c_p)$, and the on/off flag $\in\{0,1\}$. If the on/off flag is 1,  it is an ``{\it on-curve}'' point; otherwise, an ``{\it off-curve}.'' A contour is drawn to pass on-curve points. An off-curve point controls the curvature of the contour. As shown in Fig.~\ref{fig:outline}, a character contour sometimes does not have any off-curve point; in this case, the contour is drawn only with straight line segments. 
\par

\subsection{Transformer model for \Ours}
Fig.~\ref{fig:model}~(b) shows the proposed TrueType Transformer (\Ours), which performs a recognition task by using an outline data $\mathbf{f}$. \Ours is based on a Transformer encoder model, like BERT~\cite{devlin2018bert}. A sequence $\mathbf{f}$ of $N$ control points is first fed to Transformer via a single $5\times D$ fully-connected (FC) layer (depicted as a green box in (b)). To the resulting $N$\ $D$-dim vectors, another $D$-dim dummy vector, called {\em Class token}, is added. Class token is generated as a random and constant vector. 
\par
The succeeding process with Transformer is outlined as follows. The $(N+1)$ $D$-dim input vectors are fed to the multi-head self-attention module, which calculates the correlation of each pair of $(N+1)$ inputs by $M$ different self-attention modules and then outputs $(N+1)$ $D$-dim vectors by using the correlations as weights. Consequently,  $(N+1)$ input vectors are ``transformed ''  to other $(N+1)$ output vectors by the multi-head self-attention module and an FC layer, while keeping their dimensionality $D$.  This transformation operation is repeated $L$ times in Transformer, as shown in Fig.~\ref{fig:model}(b). The final classification result is given by feeding the leftmost output vector (corresponding to Class token) to a single $D\times K$ FC layer, where $K$ is the number of classes. This FC layer is the so-called MLP-head and implemented with the soft-max operation.\par
As noted above, the structure of \Ours follows BERT~\cite{devlin2018bert}, but has an important  difference in the {\em position encoding} procedure~\cite{dufter2021pe}.
In most Transformer applications, including BERT and ViT, position encoding is mandatory and affects the performance drastically~\cite{wang2021pe}. \Ours, however, does not employ position encoding. This is because the input vector $f^c_p$ already contains position information as contour ID $c$ and point ID $p$. These IDs specify the order of $N$ inputs and therefore it is possible to omit any extra position encoding. 

\section{Experiment}
\subsection{Dataset}

Google Fonts\footnote{\tt https://github.com/google/fonts} were used in our experiments.
One reason for using them is that they are annotated with five font style categories (\textit{Serif}, \textit{Sans-Serif}, \textit{Display}, \textit{Handwriting}, and \textit{Monospace}). Another reason is that Google Fonts are one of the most popular font sets for font analysis research~\cite{roy2020stefann,srivatsan2019deep,srivatsan2021scalable}.
From Google Fonts, we selected the Latin fonts that were used in STEFANN~\cite{roy2020stefann}. We then omitted \textit{Monospace} fonts because of two reasons. First, they contain only 75 fonts. Second, they are mainly characterized by not their style but their kerning rule (i.e., space between adjacent letters); therefore, some \textit{Monospace} fonts have a style of another class, such as \textit{Sans-Serif}.
Consequently, we used 456 \textit{Serif} fonts, 859 \textit{Sans-Serif}, 327 \textit{Display}, and 128 \textit{Handwriting}, respectively. Those fonts are split into three font-disjoint subsets; that is, the training set with $1,218$ fonts, the validation set with $135$, and  the test set with $471$.

\begin{figure}[t]
\begin{minipage}[b]{0.45\linewidth}
    \centering
    \includegraphics[scale=0.24,clip]{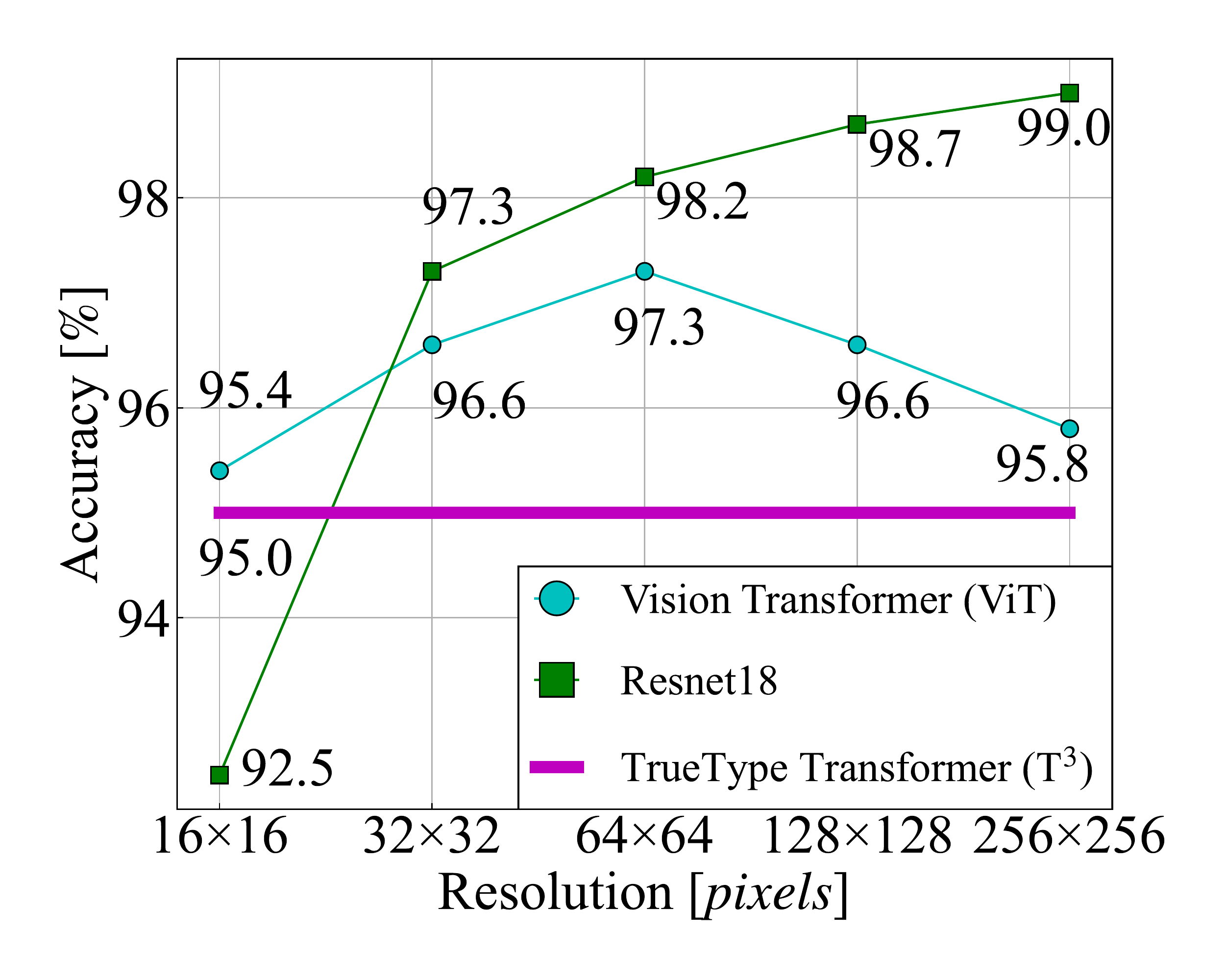}\\
    (a) Accuracy
\end{minipage}
\hspace{5mm}
\begin{minipage}[b]{0.45\linewidth}
        \centering
    \includegraphics[scale=0.12,clip]{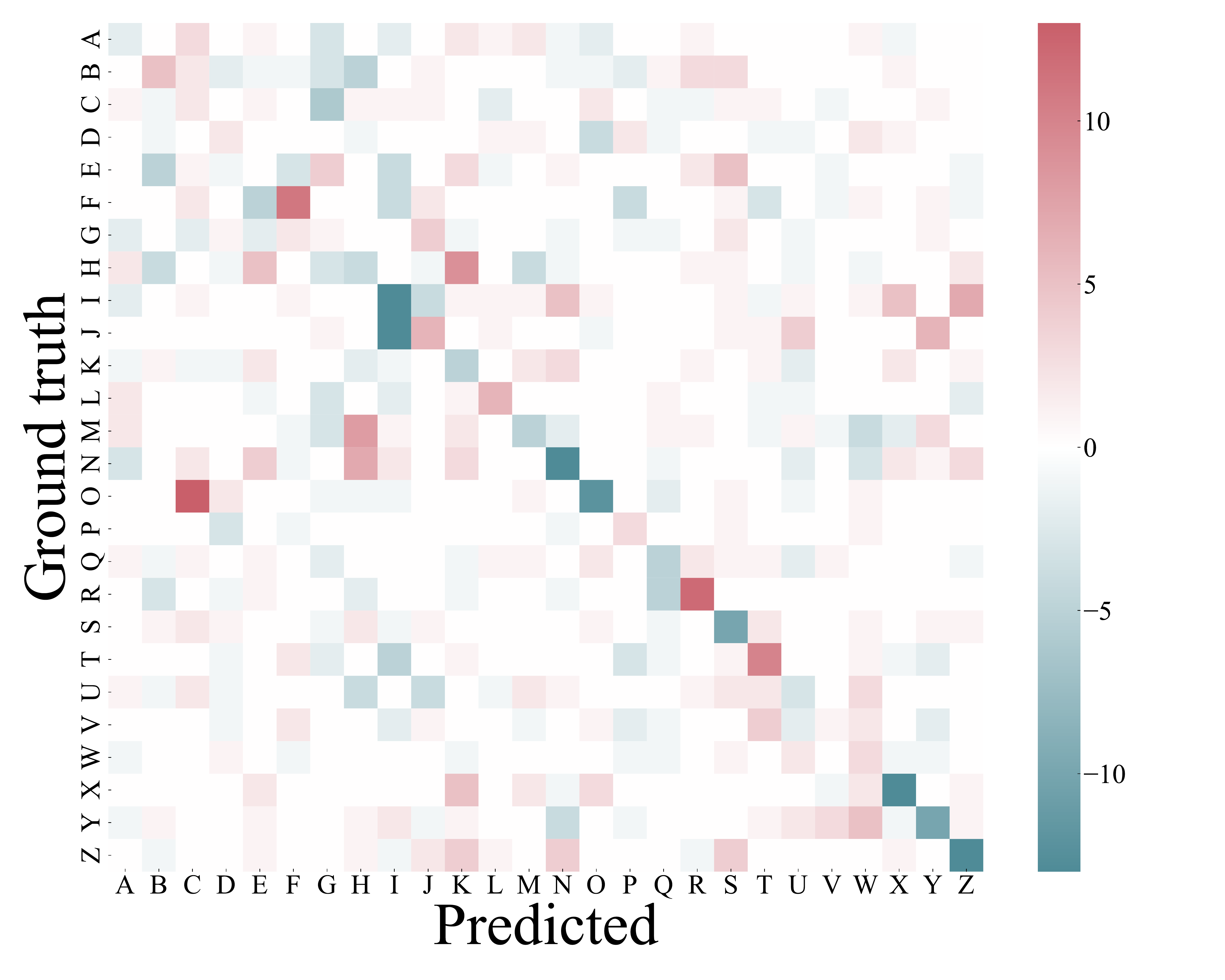}\\
    (b) Difference between confusion matrices (\Ours$-$ViT@64$\times$64)
\end{minipage}\\[-2mm]
\caption{Character recognition result. In (b), the darker green non-diagonal components and the darker red diagonal components mean that \Ours have less misrecognition.}
\label{fig:acc_char}
\bigskip

\begin{minipage}[b]{0.45\linewidth}
    \centering
    \includegraphics[scale=0.22,clip]{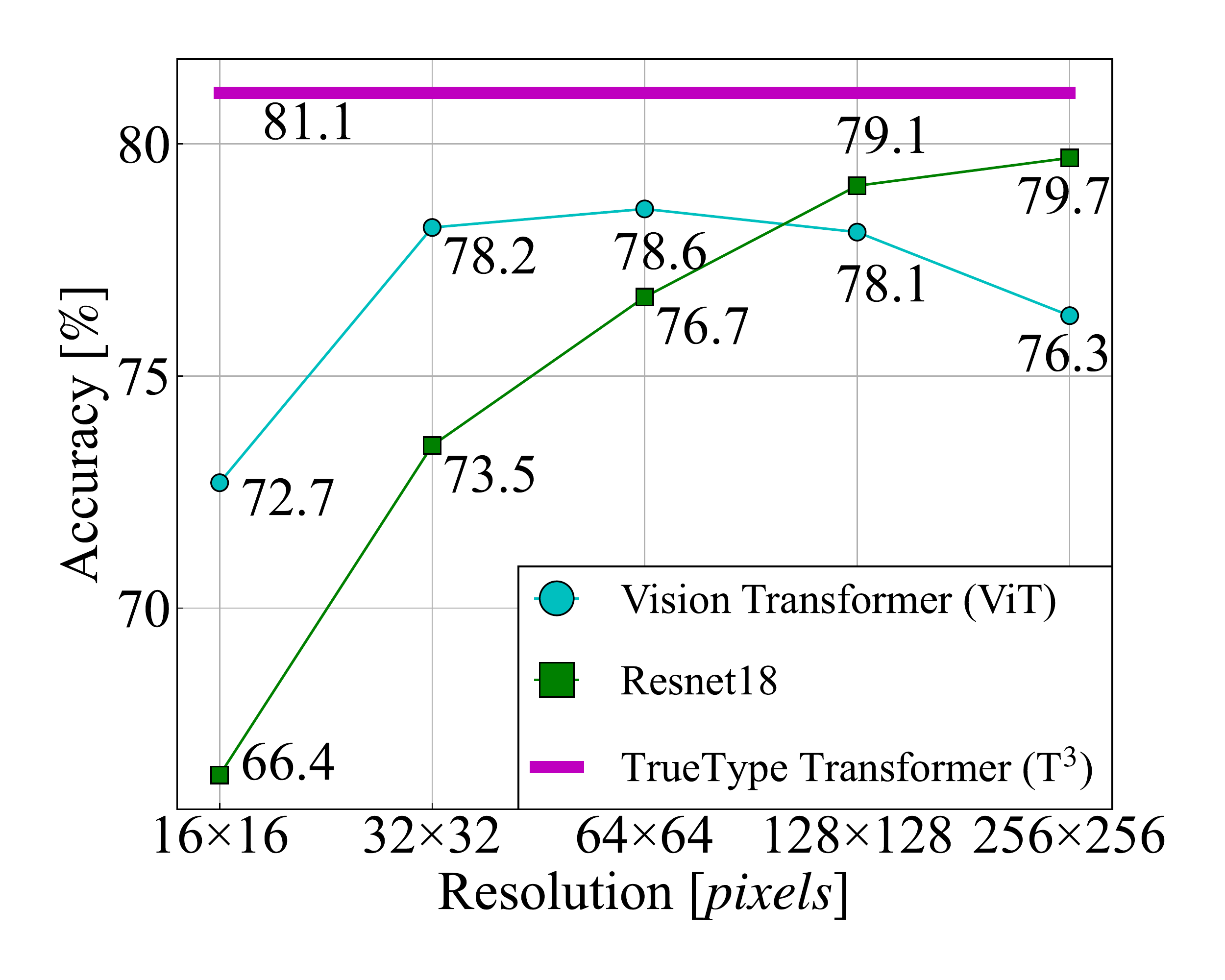}\\
    (a) Accuracy
\end{minipage}
\hspace{5mm}
\begin{minipage}[b]{0.45\linewidth}
        \centering
    \includegraphics[scale=0.22,clip]{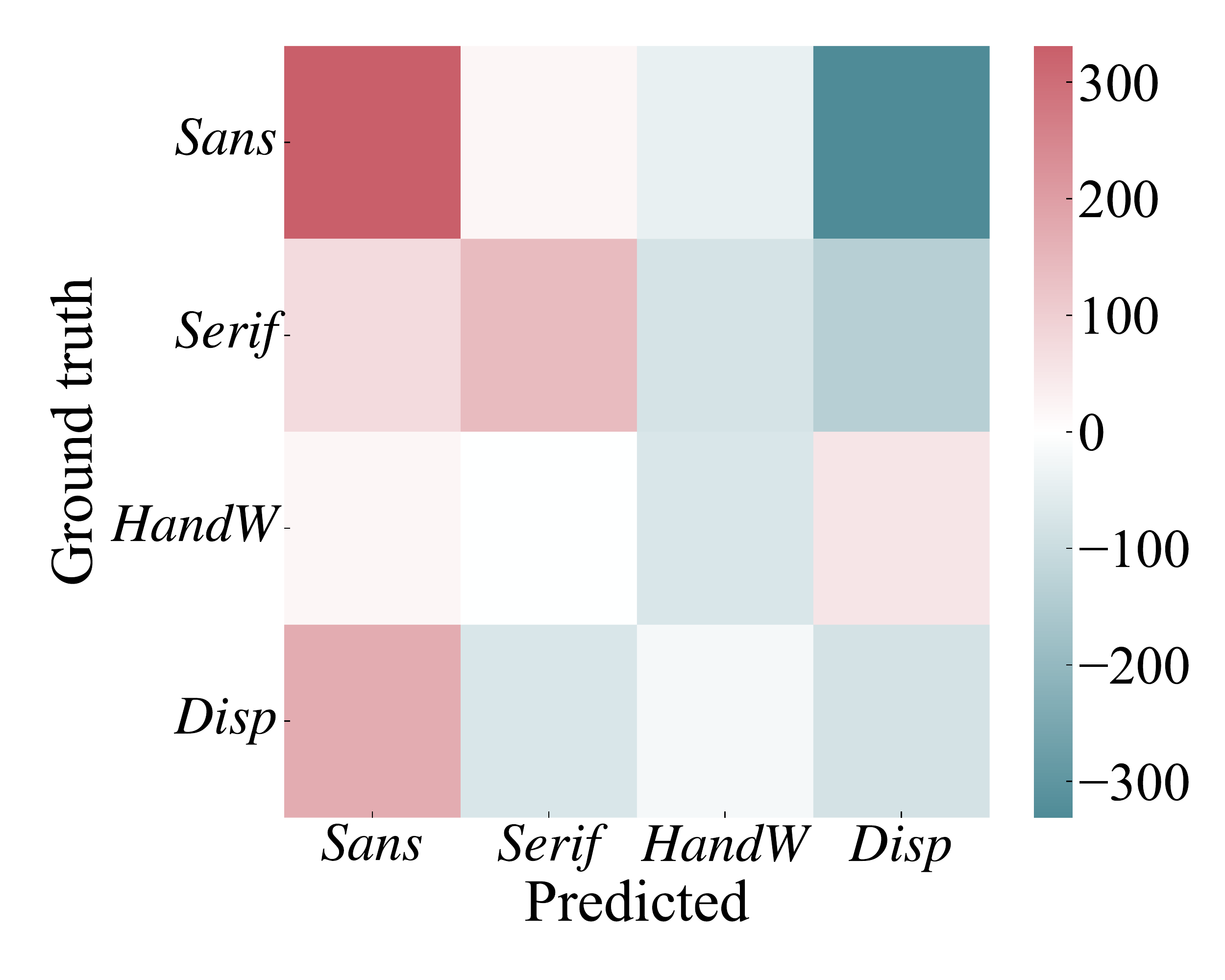}\\
    (b) Difference between confusion matrices (\Ours$-$ViT@64$\times$64)
\end{minipage}\\[-2mm]
\caption{Font style recognition result.}
\label{fig:acc_style}
\end{figure}

\subsection{Implementation details}
We set the hyperparameters of Transformer as follows. The internal dimensionality of each feature vector is set at $D=100$. The number of multiple-heads in the self-attention module is $M=5$. The number of the transformation layers was $L=6$.
As the last step of the multi-head self-attention in Transformer, each set of $M$ $D$-dimensional vectors (corresponding to each input) are converted into a single $D$-dimensional vector by a single FC layer. Consequently, the multi-head attention module accepts $(N+1)$ $D$-dimensional vectors and outputs $(N+1)$ $D$-dimensional vectors. The FC layer in the yellow box in Fig.~\ref{fig:model}~(b) converts $D$-dimensional vectors into 1024 and then $D=100$. During training,  Adam~\cite{kingma2014adam} is used as the optimizer with a learning rate of $0.0001$ to minimize the cross-entropy loss.\par

\par
As comparative methods, we used two image-based recognition methods; one is ResNet18~\cite{he2016deep}, and the other is ViT~\cite{dosovitskiy2020image}. From the outline format, we can generate bitmap images in arbitrary resolutions. We, therefore, 
examine different resolutions ($16 \times 16$, $32 \times 32$, $64 \times 64$, $128 \times 128$, and $256 \times 256$) in the later experiments. In ViT, a whole image is divided into 16 square patches; therefore, the patch size depends on the resolution. For example, for $64 \times 64$ images, each patch is $16\times 16$. Different from \Ours, ViT employs position encoding proposed in \cite{dosovitskiy2020image}. We did not use the pretrained ViT because it was pretrained with natural images from ImageNet-21k and JFT-300M instead of binary (i.e., black and white) images.

\subsection{Quantitative comparison between outline-based and image-based recognition methods}
\subsubsection{Accuracy in the character recognition task}  
Fig~\ref{fig:acc_char}~(a) shows the character recognition accuracy by \Ours, ResNet18, and ViT, at different image resolutions. Since \Ours is free from image representation, its accuracy is shown as a flat horizontal line in each graph. 
\Ours achieved $95.0\%$; although it is not higher than the image-based methods (except for ResNet@16$\times$16), it is also true that the accuracy by \Ours is not far inferior to the image-based methods. This result proves that characters can be directly recognized in their outline format. Moreover, \Ours does not need to be careful of the extra hyper-parameter, i.e., image resolution, whereas the image-based methods need\footnote{The accuracy of ResNet monotonically increases with the resolution; this is a strong advantage of ResNet. However, at the same time, the higher resolution induces more input elements. A character data in the $256 \times 256$ image format has 65,536 elements. In contrast, the outline format has only $\bar{p}\times 5$ elements, where $\bar{p}$ is the average number of the control points and about 45 in our dataset. If we use the median, $\bar{p}\sim 30$. Therefore, \Ours needs only 0.2\% elements of the $256 \times 256$ image format. Even compared with the $16\times 16$ image, \Ours needs about 60\%.}.
\par
For a more detailed comparison, Fig~\ref{fig:acc_char} (b) shows the difference of confusion matrices between \Ours and ViT@$64\times64$.
This result indicates that both methods have their own advantages. 
For example, \Ours has fewer misrecognition of `E'$\leftrightarrow$`F' and `I'$\leftrightarrow$`J.' This observation suggests that \Ours 
can capture fine differences between visually-similar classes. (This point is confirmed in the experimental results of font style recognition, which requires to discriminate visually-similar but slightly-different samples.) 
On the other hand, \Ours has more misrecognition of `H'$\to$`K' and `O'$\to$`C.'  Two letters `H' and `K' are very different in their appearance, but often not in their control points -- just removing several control points from `H,' it becomes similar to `K.' Consequently, outline and image bitmap show their own advantages, and thus their complementary usage will further improve the accuracy.\par
\par
\subsubsection{Accuracy in font style recognition task}
The advantage of \Ours is more obvious in the font style recognition task, as shown in Fig.~\ref{fig:acc_style}~(a). In this result, \Ours outperforms the image-based methods regardless of the image resolution; even with the $256 \times 256$ resolution, image-based methods could not outperform \Ours. This is because the outline format can represent fine structures of font shapes. In fact, the outline format can explicitly represent the sharpness of stroke corners, tiny serif structures, fluctuations of curvatures, etc.
\par
Fig~\ref{fig:acc_style} (b) shows the difference of confusion matrices between \Ours and ViT@$64\times64$. This result shows that the misrecognition of \textit{Sans-Serif}$\to$\textit{Display} (the darkest green element) is drastically decreased by \Ours. In fact, there are many ambiguous fonts between \textit{Sans-Serif} and \textit{Display}, and thus the style classification based on their appearance is difficult. However, in the outline format, it becomes easier to classify them. This is because, as we will see later, fonts in \textit{Display} often have many control points to realize their unique shape. Therefore, by utilizing the number of the control points for the final classification inside Transformer, \Ours can distinguish  \textit{Display} from  \textit{Sans-Serif} and vice versa.

\subsection{Qualitative comparison between outline-based and image-based recognition methods}
\subsubsection{Samples correctly recognized by \Ours}
\begin{figure}[t]
\centering
\begin{minipage}[b]{0.45\linewidth}
    \centering
    \includegraphics[width=0.85\textwidth]{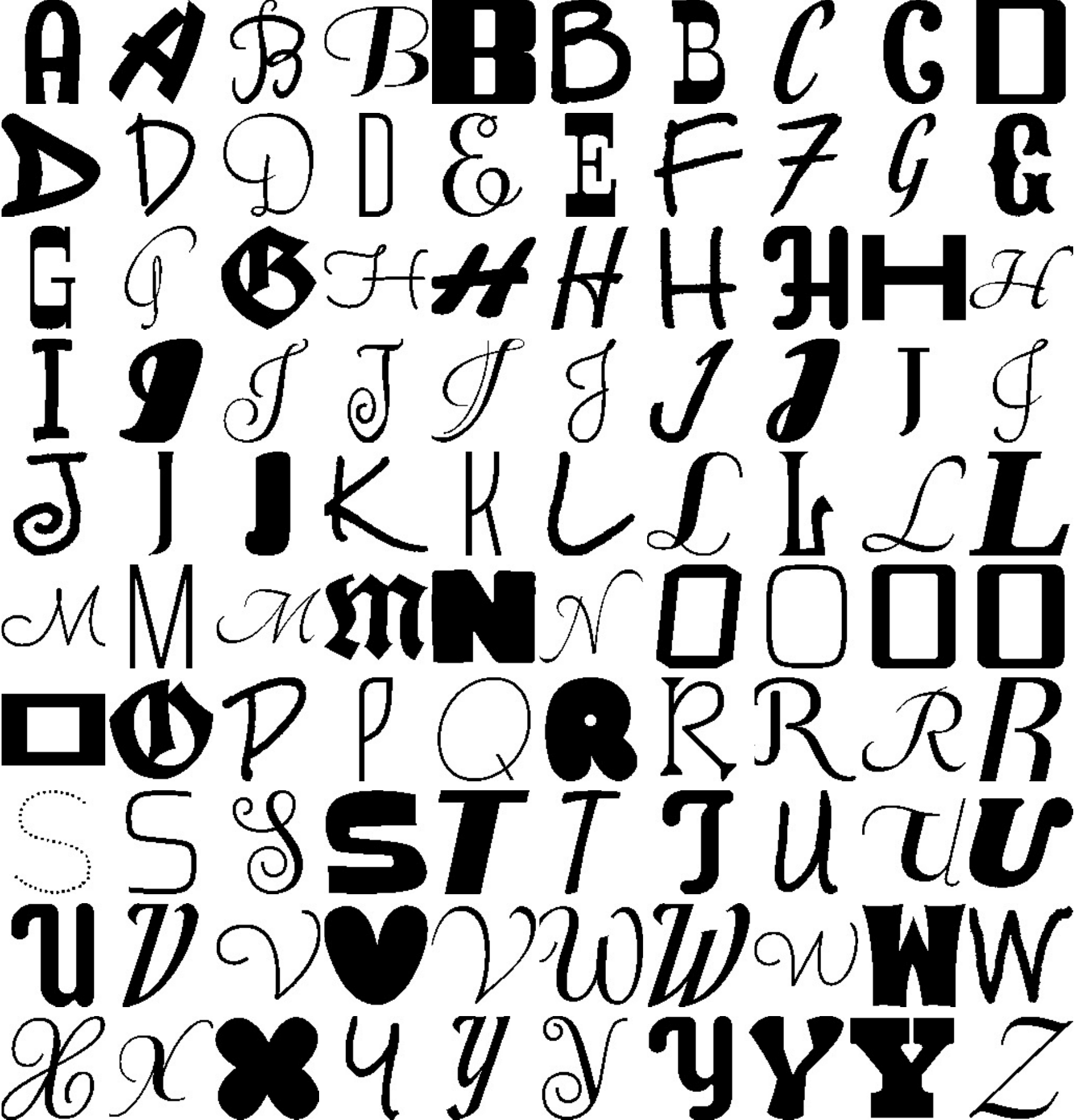}\\
    (a) Character recognition task.
    \label{fig:T3-only-class}
\end{minipage}
\hspace{0.5cm}
\begin{minipage}[b]{0.45\linewidth}
    \centering
    \includegraphics[width=0.85\textwidth]{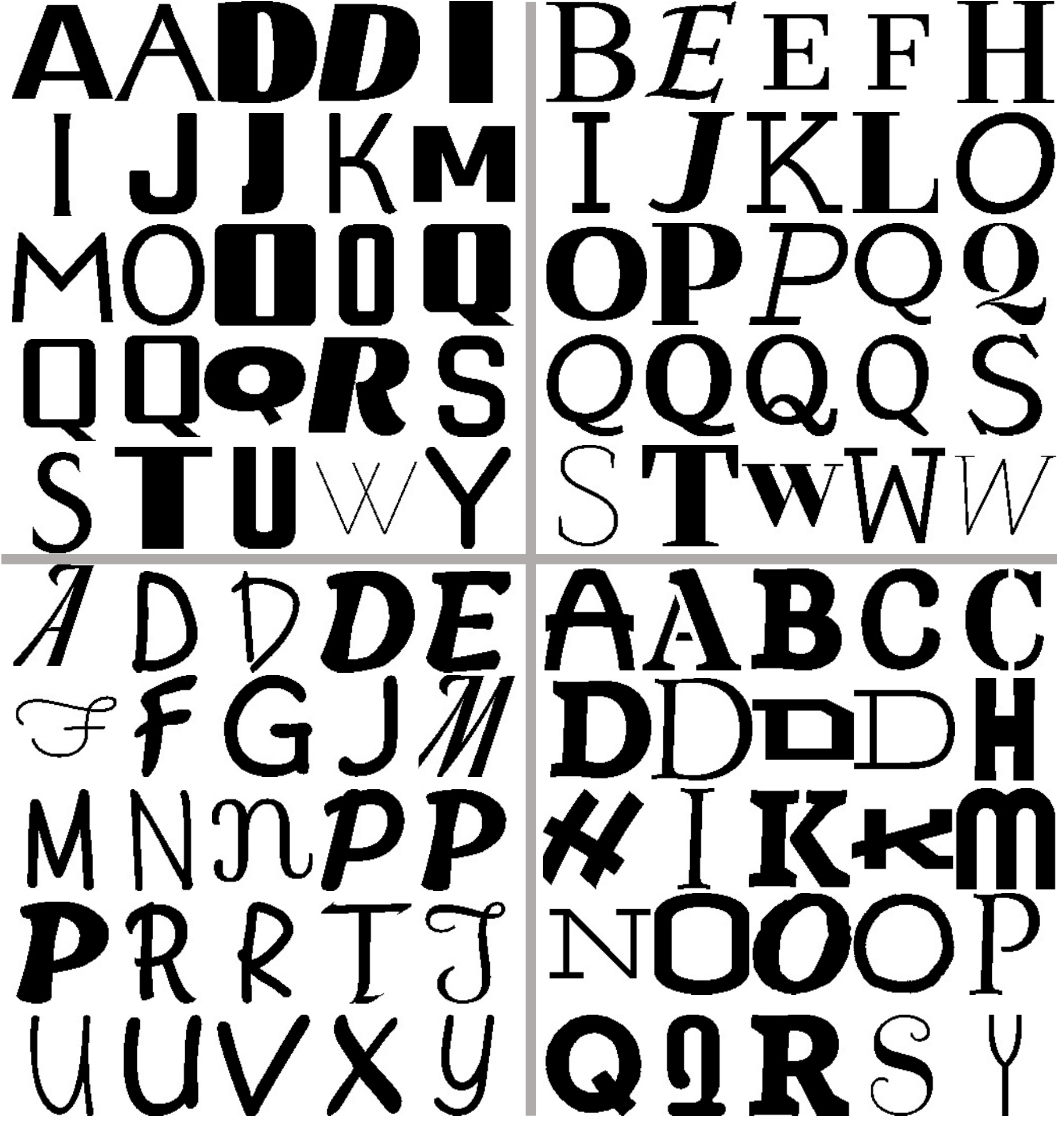}\\
    (b) Font style recognition task.
    \label{fig:T3-only-style}
\end{minipage}
\caption{Samples correctly recognized by \Ours. In (b), samples are divided into four blocks according to their style class; from top left to bottom right, \textit{Sans-Serif}, \textit{Serif}, \textit{Handwriting}, and \textit{Display}.}
\label{fig:T3-only}
\end{figure}

To understand the characteristics of \Ours via its successful results, 
we randomly selected 100 samples that were correctly recognized by \Ours and misrecognized by ResNet and/or ViT, and show them in Fig.~\ref{fig:T3-only}. In (a), we can observe a clear trend that \Ours could recognize the character class of very thin font images. 
This is reasonable because very thin strokes might be overlooked by the image-based methods, whereas they have no large difference from thicker strokes for our outline-based method. In other words, in the outline format, almost the same number of control points will be used regardless of the stroke thickness. This property realizes the robustness of \Ours to thin strokes.\par
Compared to Fig.~\ref{fig:T3-only}~(a), the samples in (b) did not show any clear trend. However, for example, the shapes of those \textit{Sans-Serif} fonts have very sharp corners and, therefore, \Ours can give the correct recognition results for them if \Ours has the ability to emphasize the sharp corner.
In Section~\ref{sec:attention-analysis}, we will conduct a more detailed analysis of the style recognition results by using the learned attention. We then can confirm that \Ours has the ability to emphasize the unique characteristics of each font style class. \par
\subsubsection{The number of control points in \textit{Display}}
For \textit{Display} style, having a large number of control points is an important clue for the correct recognition. Each character sample of \textit{Display} has 166 control points on average when its style is correctly recognized as \textit{Display}; however, 
misrecognized samples only have 47 control points on average. Fonts of \textit{Display} often have a complicated outline or a rough (i.e., non-smooth) outline and thus need to use many control points. Therefore, it is reasonable that \Ours internally uses the number of points for discriminating \textit{Display} from the others. \par

\begin{figure}[h!]
\centering
\begin{minipage}[b]{0.230\linewidth}
\centering
\includegraphics[width=0.70\textwidth]{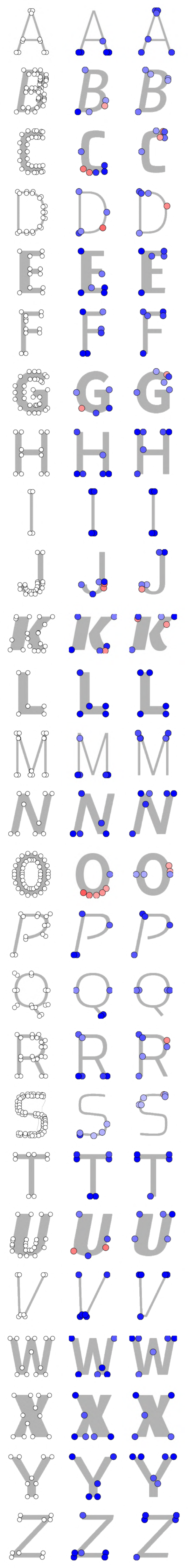}\vspace{-1mm}\\
(a)~\textit{Sans-Serif}
\end{minipage}
\smallskip
\begin{minipage}[b]{0.230\linewidth}
\centering
\includegraphics[width=0.70\textwidth]{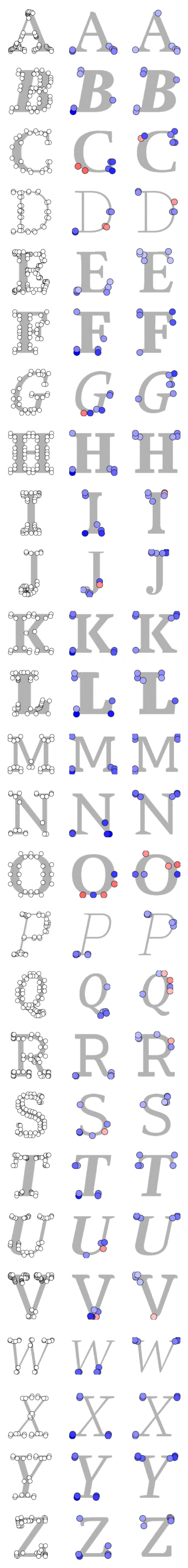}\vspace{-1mm}\\
(b)~\textit{Serif}
\end{minipage}
\smallskip
\begin{minipage}[b]{0.230\linewidth}
\centering
\includegraphics[width=0.695\textwidth]{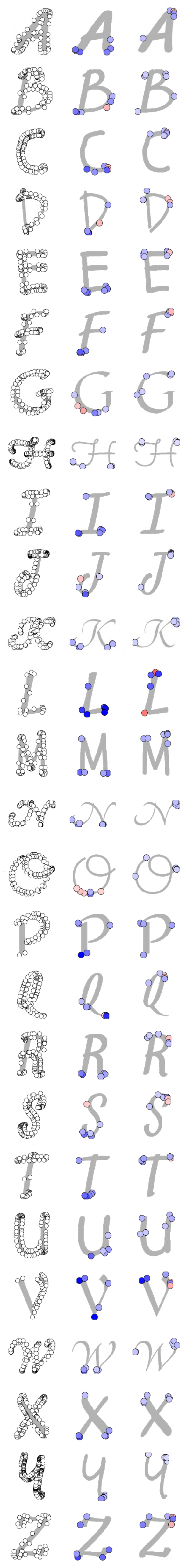}\vspace{-1mm}\\
(c)~\textit{Handwriting}
\end{minipage}
\smallskip
\begin{minipage}[b]{0.230\linewidth}
\centering
\includegraphics[width=0.70\textwidth]{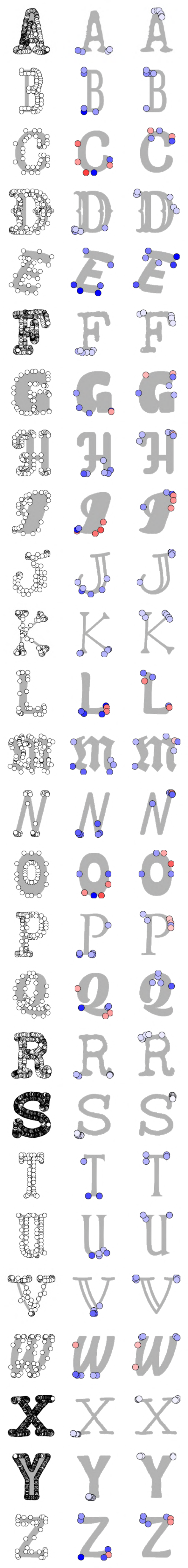}\vspace{-1mm}\\
(d)~\textit{Display}
\end{minipage}
\vspace{-0.5cm}
\caption{Visualized attention to important control points.
For each style class, the left column shows character images and their all control points. The middle and right columns plot the control points with the six largest attention in character and font style recognition, respectively. 
Since the start point and the endpoint of each contour are located at exactly the same position in the outline format, sometimes only five points appear. Blue and red points are on-curve and off-curve, respectively, and their darkness is relative to their importance, i.e., the attention weight. In each style class, 26 fonts are randomly selected and then used to print 26 characters `A'--`Z'.}
\label{fig:df_att}
\end{figure}

\subsection{Analysis of learned attention\label{sec:attention-analysis}}
\subsubsection{Visualization of learned attention}
In this section, we visualize the importance of the individual control points for the classification result, for understanding (1)~how \Ours makes its classification and (2)~where are the most important parts for each class.  For the visualization, we use learned attention. Specifically, we employ {\em attention rollout}~\cite{abnar2020quantifying} to visualize attention weight, which estimates the contribution of each input to the final classification result by using LRP (layer-wise relevance propagation)-like calculation. 
Attention rollout does not visualize the learned self-attentions (i.e., correlation among $N$ inputs) in each of $L$ transformation layers, but visualizes how each of $N$ inputs affects the final output by going through all the internal layers. \par

It should be emphasized that the point-wise visualization is an important advantage of the outline format over the image format. If we visualize important parts for image data by using XAI techniques, such as Grad-CAM, the result is somewhat blurry and fuzzy. Other XAI techniques, such as LRP, can show important pixels, but the effect of spatial smoothing in convolutional neural networks (CNN) is inevitable for them, and thus their result often becomes less reliable.  \Ours deals with a limited number (about 30) of control points, and thus the explanation of their importance becomes more reliable and sharper.\par

\subsubsection{Important points for character recognition}
As shown in the middle column in Fig.~\ref{fig:df_att}, the control points around the lower part in characters are often contributed more than the upper part to character recognition regardless of font styles. A classic typography book \cite{visibleword} claimed that the legibility of characters is lost drastically by hiding their lower part. Our result coincides with this claim. For `I,' `T,' and `Y,' the control points in the upper part are also important; however, they are still a minority compared to the lower parts.  \par

\clearpage

\subsubsection{Important points for font style recognition}
The rightmost column of each style in Fig.~\ref{fig:df_att} shows that the location of important control points is very different among the four styles. 
\begin{itemize}
    \item Important control points for \textit{Sans-Serif} show several particular trends. First, the important points are often located at the ends of straight lines. This trend reflects that the fonts of \textit{Sans-Serif} often contain sharp corners formed by two straight line segments. In other words, curves (specified by off-curve points) are not important to characterize \textit{Sans-Serif}. Second, the control points that represent a constant stroke width become important. For example, `E,' `F,' `T,' and `Z' in (a) have four important points for their top horizontal stroke; they can represent two parallel line segments, which suggest a constant stroke width. The points of `Q' also represent that the stroke width is constant at both left and right sides. The four points at the top of `H' indicate the constant stroke width in its own way. 
    \item For \textit{Serif}, the control points with larger attention locate around serifs. More interestingly, these points are not scattered but gather each other. This will be because a single point is not enough to recognize the shape of a serif. Since Transformer can deal with the correlation among the inputs, this collaborative attention among the neighboring control points was formed\footnote{When applying an XAI technique to CNN results, the importance of the neighboring pixels are often similar because of the smoothing effect of spatial convolution. In contrast, Transformer does not have any function of spatial convolution, and the relationship among all $N$ inputs is treated equally regardless of their locations. Consequently, this gathering phenomenon in Fig.~\ref{fig:df_att} proves that Transformer learns that those control points have a strong relationship to form a style.}. 
    \item  The important control points for \textit{Handwriting} have a trend similar to \textit{Serif}; they gather around the endpoint (or the corner) of strokes. This may be because to capture the smooth and round stroke endpoint shapes that are unique to \textit{Handwriting}.  
    \item \textit{Display} shows various cases (which are also observed in the other styles) because the fonts of \textit{Display} have less common trends among themselves. Sometimes a special serif shape (such as `K' in (d)) is a clue and sometimes a rough curve (such as `A' and `R' in (d)). As noted before, having a large number of control points is also an important clue to characterize complex font shapes of \textit{Display}, besides the location of important points.
\end{itemize}

\subsubsection{Comparison between two recognition tasks}
It is natural to suppose that the important points will be different in the character recognition and style recognition tasks. In the character recognition task, several control points that contribute to its legibility will have larger attention. In contrast, in the style recognition task, several control points that form decorative parts (such as serif) will have larger attention. As expected, a comparison between the middle and rightmost columns in Fig.~\ref{fig:df_att} reveals that the important points are often totally different in both tasks. For example, in the samples `A' of \textit{Handwriting} and \textit{Display} their important points locate totally upside down. \par

\subsubsection{Importance difference between on-curve and off-curve points}
\begin{figure}[t]
    \centering
    \includegraphics[width=0.70\textwidth]{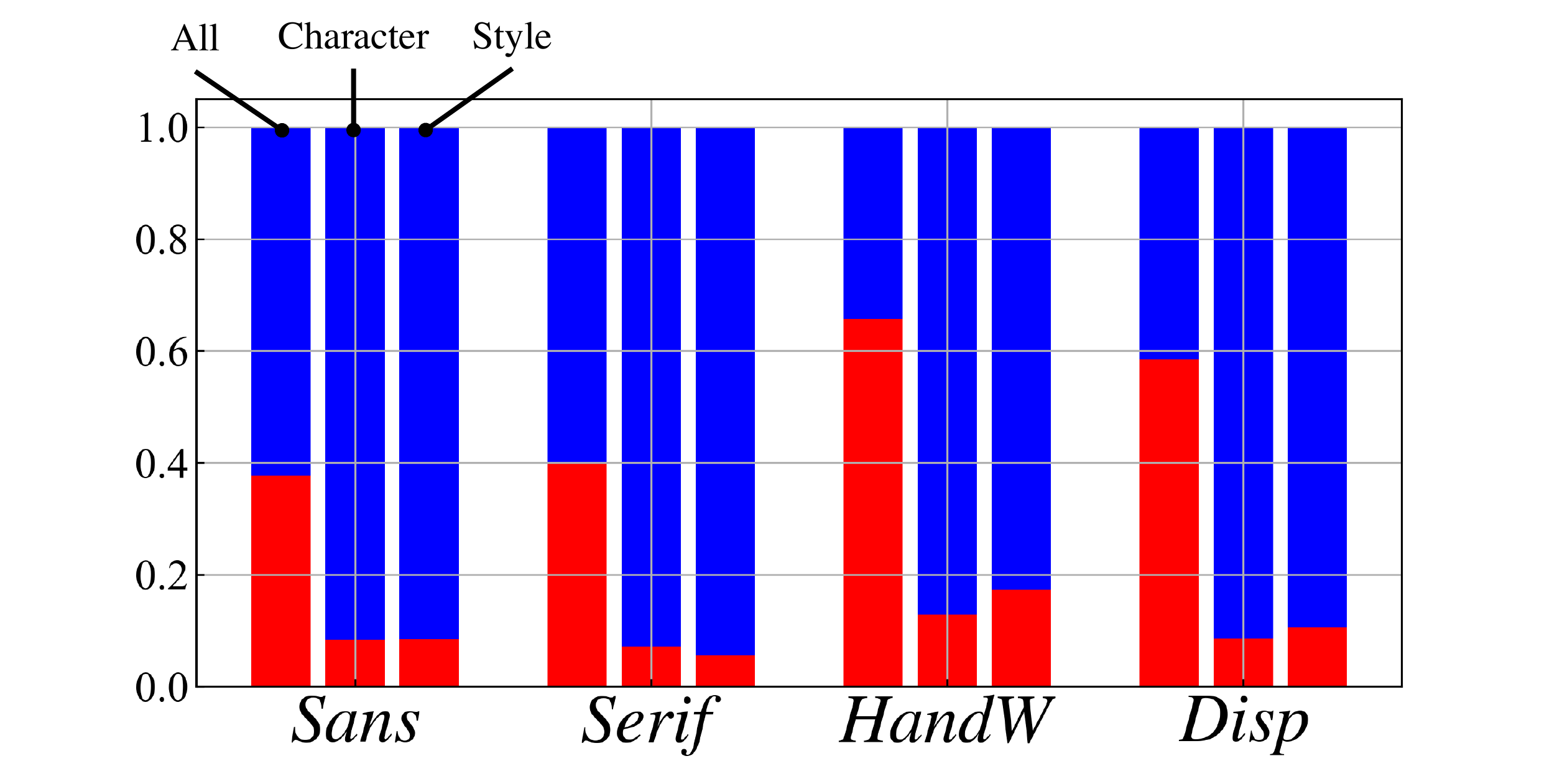}
    \caption{The ratio between on-curve points (blue) and off-curve points (red) for each style class. ``All,'' ``Character,'' and ``Style'' bars show the ratios among all control points, top six control points for character recognition, and them for font style recognition, respectively.}
    \label{fig:my_label}
\end{figure}

A careful observation of Fig.~\ref{fig:df_att} reveals that on-curve (blue) points are treated more importantly than off-curve (red) points. Except for a small number of samples, on-curve points dominate the important points with the six largest attentions in most samples. This is not because on-curve points are much more than off-curve points --- in fact, the numbers of on-curve and off-curve points do not differ significantly. 
Fig.~\ref{fig:my_label} shows the ratio between on-curve and off-curve points for each style. Specifically, the leftmost bar for each style shows the ratio, and the ratio of off-curve points is at least about 40\% (at \textit{Sans-Serif} and \textit{Serif}). For \textit{Handwriting} and\textit{Display}, its ratio is more than 50\%.\par
Fig.~\ref{fig:my_label} also shows the ratio of on-curve and off-curve among the most important six points on both recognition tasks. As observed above, the ratio of off-curve points decreases to about 10\%. This fact indicates that off-curve points are not very important for recognition tasks. For the character recognition task, this fact is convincing because the rough character shape described only by the on-curve points will be enough to recognize the character class. For the font style recognition task, however, this fact is somewhat unexpected because a fine curvature seems to be important as the clue of the style. Although the off-curve points might become more important for finer style classification tasks (such as font impression classification~\cite{Chen2019large,ueda-ICDAR2021}),  the fine curvature was not important, at least, for the current four style classes.

\section{Conclusion}
This paper reports a trial of character and font style recognition in an outline format. The outline format has different information from 
the bitmap image format, even when they seem almost identical in their appearance as images. This difference realizes several advantages of the outline format over the image format. For example, as we proved experimentally, the outline format is suitable for font style recognition since it can explicitly represent the fine structure of font shapes. 
Moreover, in the character recognition task, where the image format also has a high recognition accuracy, a comparable accuracy can be obtained with the outline format.\par
For the outline-based recognition, we developed TrueType Transformer (\Ours), which is a Transformer-based classification model. Different from an image-based classification model, such as ResNet, it can directly accept a set of an arbitrary number of control points that define the outline, i.e., the font shape. Moreover, by visualizing the attention weight learned in \Ours, we can understand which control points are important for determining the recognition results. Various analyses have been made and their results indicate the strategy of \Ours in recognizing different font styles using control points. \par
Future work will focus on the following attempts. First, we apply \Ours to a recognition task of font impressions (such as \textit{elegant} and \textit{funny}) to see its performance on this more delicate recognition task. In addition, we will observe the importance of off-curve points, which control subtle curvatures of the outline. Second, we extend \Ours 
to convert the outline data into its deformed version. In other words, we can utilize Transformer as a shape transformer. More precisely, we formulate this shape transformation task in a Transformer-based encoder-decoder system (just like a language translator~\cite{vaswani2017attention}). In these future attempts, we will be able to utilize the quantitative and qualitative analysis results given in this paper.

\bibliography{ref}
\bibliographystyle{splncs04}
\end{document}